\documentclass{article}

\usepackage{arxiv}

\usepackage[utf8]{inputenc} 
\usepackage[T1]{fontenc}    
\usepackage{url}            
\usepackage{booktabs}       
\usepackage{amsfonts}       
\usepackage{nicefrac}       
\usepackage{microtype}      
\usepackage{lipsum}		
\usepackage{graphicx}
\usepackage[numbers]{natbib}
\usepackage{doi}
\usepackage{soul}
\usepackage{multirow}
\usepackage{float}
\usepackage{nameref}
\usepackage{hyperref}
\usepackage{amsmath}
\usepackage{natbib} 
\usepackage{enumitem}
\usepackage{tabularx}
\usepackage{booktabs}
\usepackage{hyperref}       
\title{Neuroevolution of Physics-Informed Neural Nets: Benchmark Problems and Comparative Results}
\author{
  \href{https://orcid.org/0000-0003-1184-8608}{\includegraphics[scale=0.06]{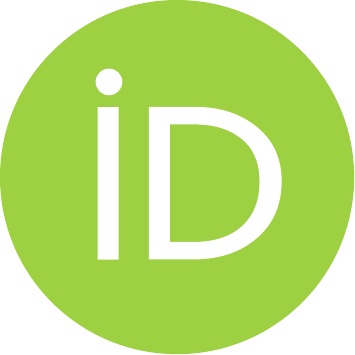}}\hspace{1mm}Nicholas Wei Yong Sung\\
  Center for Frontier AI Research,\\
  Agency for Science, Technology, and Research,\\
  1 Fusionopolis Way, \#16-16 Connexis\\
  Singapore, S138632\\
  \texttt{sungnwy@cfar.a-star.edu.sg}
  \and
  \href{https://orcid.org/0000-0002-3215-1888}{\includegraphics[scale=0.06]{orcid.pdf}}\hspace{1mm}Jian Cheng Wong\\
  Institute for High Performance Computing,\\
  Agency for Science, Technology, and Research,\\
  and Nanyang Technological University\\
  1 Fusionopolis Way, \#16-16 Connexis\\
  Singapore, S138632\\
  \texttt{wongj@ihpc.a-star.edu.sg}
  \and
  \href{https://orcid.org/0000-0003-4813-4529}{\includegraphics[scale=0.06]{orcid.pdf}}\hspace{1mm}Chin Chun Ooi\\
  Center for Frontier AI Research,\\
  Agency for Science, Technology, and Research,\\
  1 Fusionopolis Way, \#16-16 Connexis\\
  Singapore, S138632\\
  \texttt{ooicc@cfar.a-star.edu.sg}
  \and
  \href{https://orcid.org/0000-0002-6080-855X}{\includegraphics[scale=0.06]{orcid.pdf}}\hspace{1mm}Abhishek Gupta\\
  Singapore Institute of Manufacturing Technology,\\
  Agency for Science, Technology, and Research,\\
  2 Fusionopolis Way, \#08-04 Innovis\\
  Singapore, S138634\\
  \texttt{\string abhishek\textunderscore gupta@simtech.a-star.edu.sg}
  \and
  \href{https://orcid.org/0000-0002-5626-9688}{\includegraphics[scale=0.06]{orcid.pdf}}\hspace{1mm}Pao-Hsiung Chiu\\
  Institute for High Performance Computing,\\
  Agency for Science, Technology, and Research,\\
  1 Fusionopolis Way, \#16-16 Connexis\\
  Singapore, S138632\\
  \texttt{chiuph@ihpc.a-star.edu.sg}
  \and
  \href{https://orcid.org/0000-0002-4480-169X}{\includegraphics[scale=0.06]{orcid.pdf}}\hspace{1mm}Yew-Soon Ong\\
  Agency for Science, Technology, and Research,\\
  and Nanyang Technological University\\
  1 Fusionopolis Way, \#16-16 Connexis\\
  Singapore, S138632\\
  \texttt{asysong@ntu.edu.sg}
    }

\begin{document}
\maketitle

\begin{abstract}
  The potential of learned models for fundamental scientific research and discovery is drawing increasing attention worldwide. Physics-informed neural networks (PINNs), where the loss function directly embeds governing equations of scientific phenomena, is one of the key techniques at the forefront of recent advances. PINNs are typically trained using stochastic gradient descent methods, akin to their deep learning counterparts. However, analysis in this paper shows that PINNs' unique loss formulations lead to a high degree of complexity and ruggedness that may not be conducive for gradient descent. Unlike in standard deep learning, PINN training requires globally optimum parameter values that satisfy physical laws as closely as possible. Spurious local optimum, indicative of erroneous physics, must be avoided. Hence, neuroevolution algorithms, with their superior global search capacity, may be a better choice for PINNs relative to gradient descent methods. Here, we propose a set of five benchmark problems, with open-source codes, spanning diverse physical phenomena for novel neuroevolution algorithm development. Using this, we compare two neuroevolution algorithms against the commonly used stochastic gradient descent, and our baseline results support the claim that neuroevolution can surpass gradient descent, ensuring better physics compliance in the predicted outputs. 
\end{abstract}


\section{Introduction}
A physics-informed neural network (PINN) is a type of deep learning model that incorporates governing physical equations and constraints into the training process. PINNs combine the expressive power of neural networks with the guarantees accorded by compliance with known physics principles, ensuring that the networks' predictions remain consistent with the fundamental laws of nature \cite{raissi2019physics}.  The utility of this framework has been demonstrated on various complex scientific and engineering problems, towards purposes such as predicting the behaviour of physical systems, modelling the properties of materials, and assimilating data from experiments and simulations \cite{karniadakis2021physics}.

While promising and very effective when successfully learned, PINN training has been shown to be complex and prone to various failure mechanisms even when state-of-the-art stochastic gradient descent-based (SGD-based) methods are used for training \cite{krishnapriyan2021characterizing}. A variety of challenges have been identified in previous work, such as the need for proper balancing of competing loss terms in a PINN’s objective function \cite{de2022mopinns}, the impact of learning biases \cite{rahaman2019spectral}, sensitivity to initialization \cite{wong2022learning} and a consequent inability to escape spurious local minima \cite{garipov2018loss,rohrhofer2021pareto,gopakumar2022loss}. Critically, these prior work typically use local optimization methods like gradient descent. Hence, this paper first seeks to provide insights into key characteristics of the loss function landscape to explain some of these challenges. Based on the observed spurious local minima traps, we then make a case for the evaluation of global search methods like \textit{evolutionary algorithms} in place of local methods like SGD for PINN training. It is worth emphasizing that the goal in PINNs is to arrive precisely at a globally optimum solution that satisfies physics as closely as possible. An inferior local optimum indicating erroneous physics must be avoided.  

Evolutionary algorithms (EAs) form a class of artificial intelligence methods that are commonly used to approximate solutions to non-convex optimization problems with multiple local optima. They can explore large spaces of possible solutions effectively even for complex or nonlinear systems \cite{morse2016simple}, with their application in the context of neural nets referred to as \textit{neuroevolution} \cite{stanley2019designing}. In PINNs, neuroevolution can be used to find a near globally optimal configuration of the network's parameters that model a given physical system  accurately, and hence may present certain advantages over conventional gradient descent methodologies in view of their complicated optimization landscapes \cite{wong2021can}. However, despite the potential benefits of neuroevolution, there has been relatively little research in this area. With that in mind, this work aims to systematically evaluate the effectiveness of neuroevolution relative to more commonly used SGD optimizers for PINN traning. 

Further, to facilitate systematic and consistent evaluation of the performance and speed of EAs, we propose a set of five PINN benchmark problems which are representative of diverse real-world phenomena across classical mechanics, heat and mass transfer, fluid dynamics and wave propagation (e.g. acoustics), and open-source the codes and ground truth solutions. Specifically, we demonstrate the use of this benchmark suite for evaluating neuroevolutionary algorithms such as the widely adopted Covariance Matrix Adaptation Evolution Strategy (CMA-ES) \cite{hansen2006cma} and exponential Natural Evolution Strategies (xNES) \cite{wierstra2014natural}. The loss and training time of EAs and SGD are reported and compared for these five benchmark problems using a JAX-based implementation \cite{tang2022evojax,abadi2016tensorflow}. We then objectively assess the relative efficacy and efficiency of EAs and SGD for training PINNs. Notably, neuroevolution consistently provides competitive or superior convergence performance. These benchmarks and baseline results are provided to motivate the development and testing of future state-of-the-art neuroevolutionary algorithms, especially those crafted for application to PINNs. Implementations of all benchmark problems, including analytical and simulation solutions, are provided on GitHub to aid community development (\url{https://github.com/nicholassung97/Neuroevolution-of-PINNs}). 

The remainder of the paper is organized as follows. Section \ref{Neuro-Potential} presents a landscape study of the PINN loss, outlining the potential and capacity of neuroevolution to address challenges faced by gradient-based learning in the PINN context. Section 3 describes the optimization algorithms and deep learning frameworks used. Sections 4 and 5 introduce the five benchmark problems and performance metrics, respectively. Section 6 evaluates the performance and speed of two neuroevolution algorithms against SGD across the benchmark problems. Finally, Section 7 presents concluding remarks and direction for future research.

\section{Positioning Neuroevolution for PINNs} \label{Neuro-Potential}
In this section, we first present the problem setup for PINNs, contrasting it to standard deep neural networks (DNNs). The loss landscapes of the two models are analyzed, making the case for neuroevolution as a strong candidate for PINN training.

\subsection{PINN vs DNN Loss Functions} 
\label{PINN-v-DNN}
A typical PINN uses a multilayer perceptron (MLP) representation, $u(x,t;\boldsymbol{w})$, to model the dynamic scalar quantity of a physical system ($u$) in space, $x\in\Omega$, and time, $t\in[0,T]$. The network parameters, $\boldsymbol{w}$, are optimized for this purpose. Additionally, $u$ must adhere to known mathematical constraints, such as partial differential equations (PDEs) of the general form:
\begin{subequations} \label{eq:pde_ibc_eqn}
\begin{align}
&\mathcal{N}_t[\boldsymbol{u}(x,t)] + \mathcal{N}_x[\boldsymbol{u}(x,t)] = 0, &\quad\quad x\in\Omega, t\in[0,T] \label{eq:pde_eqn} \\
&\boldsymbol{u}(x,t=0) = \boldsymbol{u}_0(x), &\quad x\in\Omega \label{eq:ic_eqn} \\
&\mathcal{B}[\boldsymbol{u}(x,t)] = g(x,t), &\quad x\in\partial\Omega, t\in[0,T] \label{eq:bc_eqn}
\end{align}
\end{subequations} 
The temporal derivative is represented by $\mathcal{N}_t[\cdot]$ and the general nonlinear differential operator $\mathcal{N}_x[\cdot]$ can include any combination of nonlinear terms of spatial derivatives. The initial state of the system at time $t=0$ is given by $u_0(x)$, and the boundary operator $\mathcal{B}[\cdot]$ ensures that the desired boundary condition $g(x,t)$ is satisfied at the domain boundary ($\partial\Omega$). $\mathcal{B}[\cdot]$ can be either an identity operator or a differential operator.

Hence, the loss function of a PINN is defined as:
\begin{subequations} \label{eq:physics_loss_fn}
\begin{align}
L_{PINN} &= \lambda_{PDE} L_{PDE} + \lambda_{IC} L_{IC} + \lambda_{BC} L_{BC} \\
L_{PDE} &= \lVert \mathcal{N}_t[\boldsymbol{u}(\cdot;\boldsymbol{w})] + \mathcal{N}_x[\boldsymbol{u}(\cdot;\boldsymbol{w})] \rVert _{L^2(\Omega \times [0,T])}^2 \\
L_{IC} &= \lVert \boldsymbol{u}(\cdot,t=0;\boldsymbol{w}) - \boldsymbol{u}_0(\cdot) \rVert _{L^2(\Omega)}^2 \\
L_{BC} &= \lVert \mathcal{B}[\boldsymbol{u}(\cdot;\boldsymbol{w})] - g(\cdot) \rVert _{L^2(\partial\Omega \times [0,T])}^2
\end{align}
\end{subequations} 
The PINN loss sums the PDE residuals across the entire spatio-temporal domain ($\Omega \times [0,T]$), and the mean squared error computed against pre-specified initial condition(s) and boundary condition(s). Although the PINN loss is formulated over the entire continuous domain, for practical purposes, the residuals are usually determined only at a finite set of $n$ collocation points $D={(x_i,t_i )}_{i=1}^n$ which correspond to spatio-temporal points in the domain. 

In contrast to PINNs where there is no need for labelled training data, the loss function of a standard DNN computes the mean squared error between the DNN output $\hat{u}$ against the target $u$ over $n$ labelled collocation points:
\begin{equation} 
\label{eq:dnn_eqn}
L_{DNN} = \lVert u_i - \hat{u}_i \rVert _{L^2(D)}^2
\end{equation}

\subsection{Complexity of PINN Loss Landscapes}\label{loss-complexity}

As is clear from Section~\ref{PINN-v-DNN}, loss functions used in PINNs and DNNs differ significantly. Hence, we chose a single instantiation of the one-dimensional convection-diffusion equation (defined by $vu_x=u_{xx}$), and trained both a DNN and PINN for further analysis. As we increase the velocity scalar ($v$), the problem becomes more difficult to solve due to a steeper gradient in $u$. We used $v = 6$ as it was moderately difficult to solve. Both networks were trained with the same optimizer, collocation points, boundary points and network architecture. 
We use a 3-layer fully-connected neural network with 10 neurons per layer, a hyperbolic tangent activation function, and randomly sample collocation points ($x$, $t$) on the domain. We then projected the training loss values in the plane spanned by the first two principal Hessian directions as per previously reported literature to get an intuition for the loss landscape \cite{prendergast2007sensitivity}. 

\begin{figure}[htbp]
\begin{center}
\centerline{\includegraphics[width=1\linewidth]{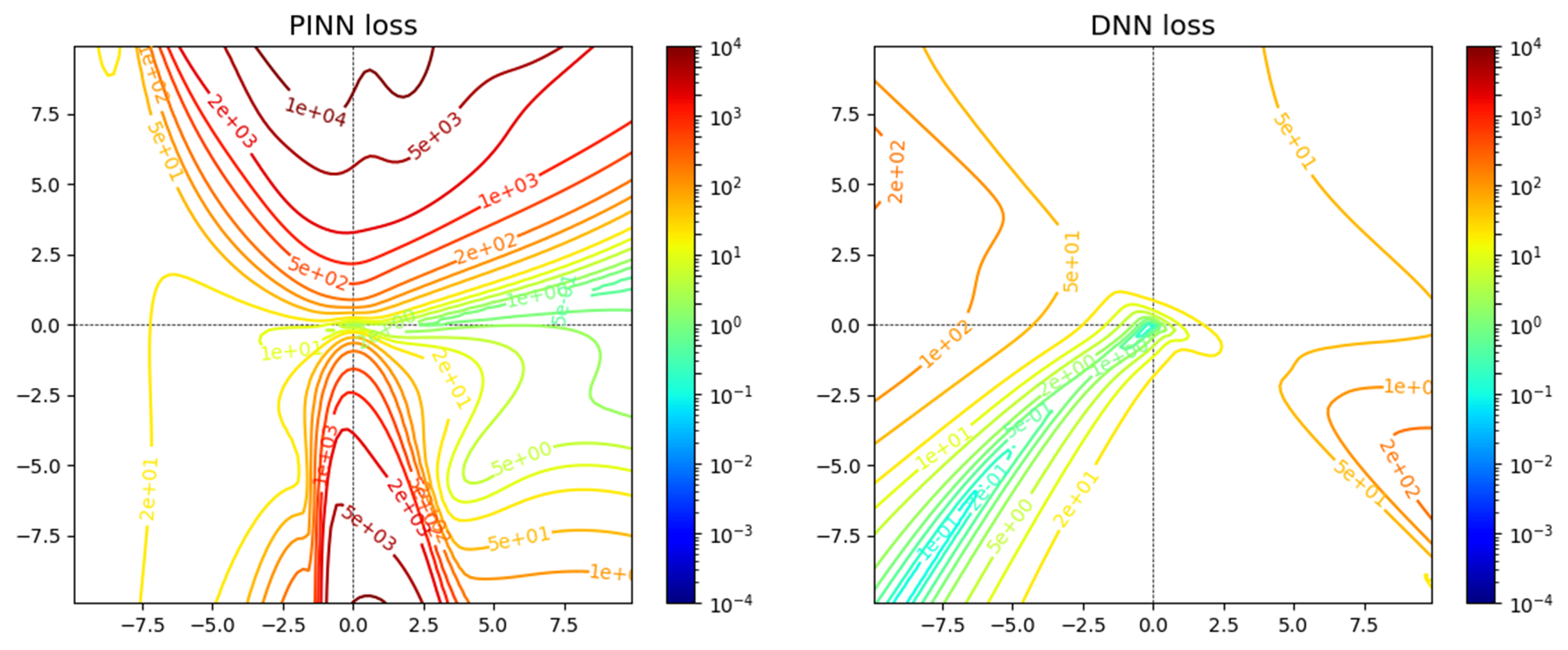}}
\caption{The plots contrast the local loss landscapes of the PINN and the DNN after 100,000 training iterations on the convection-diffusion equation $6u_x = u_{xx}$. The losses are plotted along the first two principal Hessian directions.}
\label{fig:loss-landscape-full}
\end{center}
\vspace*{-7mm}
\end{figure}

After 100,000 training iterations, we observe a much higher degree of complexity and ruggedness of the loss landscape for PINN compared to DNN, as illustrated in Fig.~\ref{fig:loss-landscape-full}. These loss landscapes provide evidence that the loss function in PINN is indeed significantly more complex, thereby raising questions about the suitability of conventional gradient-based learning algorithms. While this is a specific example of the increased complexity due to the PINN loss, similar observations have also been observed for other differential equations ~\cite{krishnapriyan2021characterizing}. 

\subsection{Analysing the Loss Landscapes at Model Initialization} \label{loss-initialization}

In addition to the rugged loss landscape that a PINN has to navigate during training, Wong et al. \cite{wong2022learning} have previously also shown that PINNs are susceptible to being trapped in a deceptive local minima at the point of model initialization. The following proposition from \cite{wong2022learning} highlights this point.

\textit{Proposition 1: Let $\boldsymbol{u}(\boldsymbol{x};\boldsymbol{w})$ be a PINN with $L$ fully connected layers, $n$ neurons per layer, activation function $f=tanh$, and parameters $\boldsymbol{w}$. Let all the dense layers be initialized by the Xavier method for network weights, i.e., $w_{l}$’s are i.i.d. $\mathcal{N}(0,\frac{1}{fan_{in}+ fan_{out}})$, where $fan_{in}$ and $fan_{out}$t are the number of inputs and outputs for the dense layer. Then, as $n\to\infty$, $\boldsymbol{u}(\boldsymbol{x};\boldsymbol{w})$ trivially satisfies arbitrary differential equations of the form $F(\frac{\partial \boldsymbol{u}}{\partial \boldsymbol{x}},\frac{\partial^{2} \boldsymbol{u}}{\partial^{2} \boldsymbol{x}},...,\frac{\partial^{k} \boldsymbol{u}}{\partial^{k} \boldsymbol{x}})=0$ where $F(\varphi_{1},\varphi_{2},\varphi_{3},...)= \sum_{i} a_{i}\varphi_{i} + \sum_{i\leq j} b_{ij}\varphi_{i}\varphi_{j} + \sum_{i\leq j\leq k} c_{ijk}\varphi_{i}\varphi_{j}\varphi_{k} +...$ with probability 1 at initialization.}

Briefly, Proposition 1 shows that a PINN tends to prematurely satisfy certain (commonly occurring) PDEs at initialization without accounting for boundary conditions. This hints at a deceptive local minimum of the overall PINN loss at initialization, which has zero PDE loss but violates the initial or boundary conditions. While proven in the limit of infinitely wide layers, we further empirically corroborate this observation in the context of finite-width PINNs. 

\begin{figure}[htbp]
\begin{center}
\centerline{\includegraphics[width=1\linewidth]{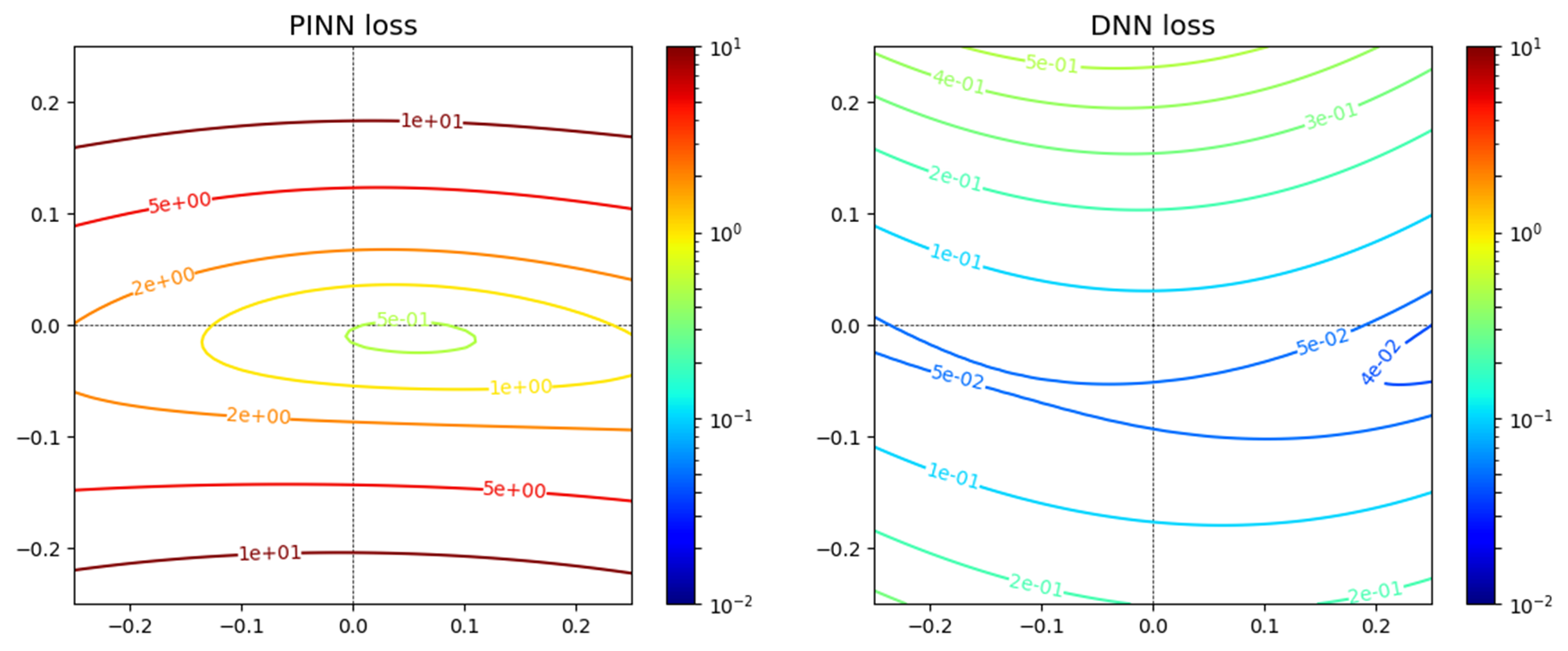}}
\caption{The plots show a zoomed-in view of the local loss landscapes of the PINN and the DNN at the point of initialization (with Xavier method). The losses are plotted along the first two principal Hessian directions.}
\label{fig:loss-landscape-initial}
\end{center}
\vspace*{-5mm}
\end{figure}

\begin{figure}[htbp]
\begin{center}
\centerline{\includegraphics[width=1\linewidth]{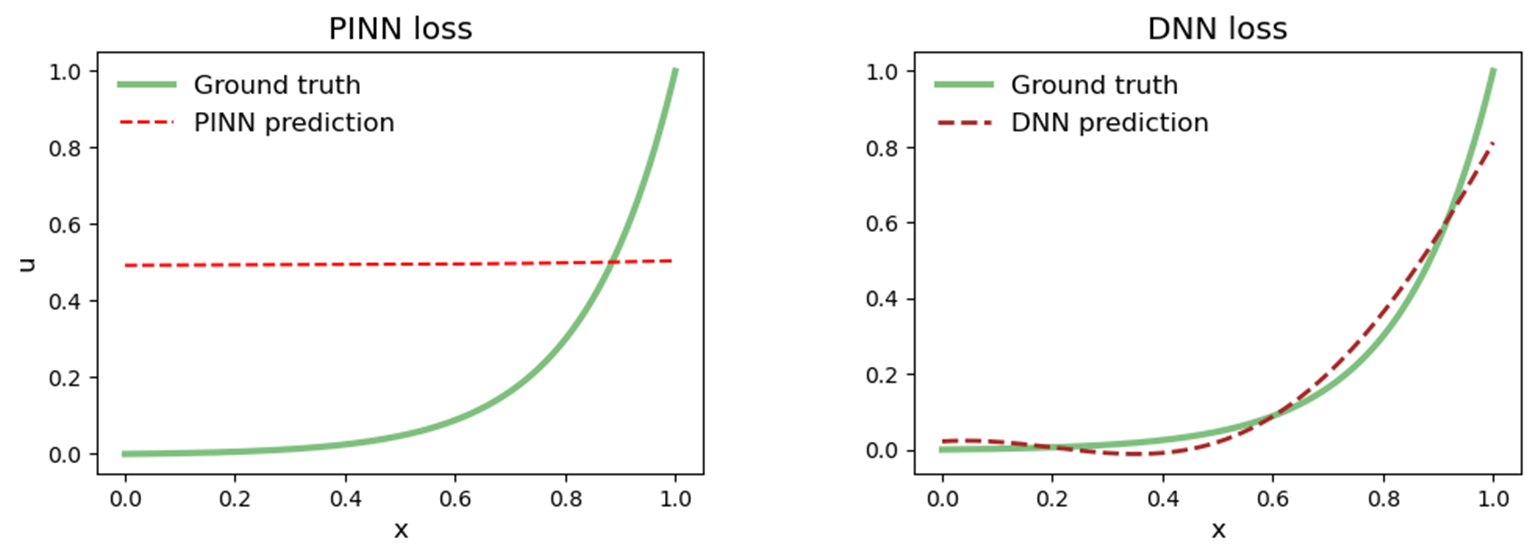}}
\caption{Plots of the solution obtained by the PINN and the DNN after 1,000 training iterations for the convection-diffusion differential equation $6u_x = u_{xx}$.}
\label{fig:prediction-1k}
\end{center}
\vspace*{-5mm}
\end{figure}

Using the same problem set-up as in Section \ref{loss-complexity}, the loss landscapes of the PINN and DNN under the same initialization are visualized in Fig.~\ref{fig:loss-landscape-initial} along the first two principal Hessian directions. The first two principal eigenvalues along those directions were also computed, and found to be 509.6 and 35.7 for the PINN, and 9.9 and -1.5 for the DNN. The two large positive eigenvalues for PINN suggest that the PINN at initialization is indeed near a local minimum, as is consistent with Proposition 1. In contrast, the presence of a negative eigenvalue for DNN indicates that there are descent paths available for a gradient-based learning algorithm to start moving towards a potential solution. Indeed, when we plot the PINN and DNN predicted outputs after 1,000 training iterations (Fig.~\ref{fig:prediction-1k}), we observed that the DNN can already approximate the ground truth reasonably well, whereas the PINN model is still far from the solution. The flatness of the PINN output suggests that this local minimum only minimizes the PDE residual terms but not the boundary conditions as is required for the true solution. This further provides evidence that the PINN is trapped in the vicinity of a deceptive local minimum at initialization. 

Clearly, PINNs initialize near spurious local minima that only satisfy one of the multiple loss terms, thus posing difficulties for optimization using local optimization methods like SGD. Conversely, the lack of a nearby, deleterious local minimum at initialization makes DNN training more amenable to optimization using SGD. Hence, global optimization methods like neuroevolution are promising for training PINNs due to their greater capability for exploring the loss landscape and escaping local minima.

\begin{table*}[t]
\footnotesize
\centering
\caption{Detailed definitions of PDEs and their specific instantiations as used in this set of benchmark problems.}

\begin{tabular}{llll}\toprule
    \textbf{Problem / PDE} & \textbf{PDE Parameter} & \vtop{\hbox{\strut \textbf{Computational}}\hbox{\strut \textbf{Domain}}} & \vtop{\hbox{\strut \textbf{Initial/Boundary}}\hbox{\strut \textbf{Condition}}}  \\ \hline
    \vtop{\hbox{\strut Convection-diffusion:}\hbox{\strut $vu_x - ku_{xx} = 0$}} & \vtop{\hbox{\strut $v = 6$}\hbox{\strut $k = 1$}} & $x\in[0,1]$ & $u(x = 0) = 0$ ; $u(x = 1) = 1$ \\ 
    \hline
    \vtop{\hbox{\strut Projectile motion:}\hbox{\strut $x_{tt} = 0$}\hbox{\strut $y_{tt} + g = 0$}}  & $g = 3.7$ &  $t\in[0,5.5]$ & \vtop{\hbox{\strut $x(t=0) = 0$ ; $y(t=0) = 2$}\hbox{\strut $x_t(t=0) = V_0cos(\frac{\alpha_0\pi}{180})$ ; $y_t(t=0) = V_0sin(\frac{\alpha_0\pi}{180})$}\hbox{\strut $V_0 = 10, \alpha_0 = 80^{o}$}} \\ 
    \hline
    \vtop{\hbox{\strut Korteweg–De Vries (KdV):}\hbox{\strut $u_t + v_1uu_x  + v_2u_{xxx} = 0$}} & \vtop{\hbox{\strut $v_1 = 1$}\hbox{\strut  $v_2 = 0.001$}} & \vtop{\hbox{\strut $x\in[0,1.5]$}\hbox{\strut  $t\in[0,2]$}} & \vtop{\hbox{\strut $u(x,t = 0) = \frac{3c_1}{(cosh(a_1(x-x_1)))^2} + \frac{3c_2}{(cosh(a_2(x-x_2)))^2}$}\hbox{\strut $a_1 = \frac{1}{2}\sqrt{\frac{c_1}{v_2}}, a_2 = \frac{1}{2}\sqrt{\frac{c_2}{v_2}}$}\hbox{\strut $c_1 = 0.3, c_2 = 0.1, x_1 = 0.4, x_2 = 0.8$}} \\ 
    \hline
    \vtop{\hbox{\strut Linearized Burgers:}\hbox{\strut $u_t + v_1u_x - v_2u_{xx} = 0$}} & \vtop{\hbox{\strut $v_1 = 1$}\hbox{\strut  $v_2 = 0.02$}} & \vtop{\hbox{\strut $x\in[-1.5,4.5]$}\hbox{\strut  $t\in[0,2]$}} & \vtop{\hbox{\strut $u(x,t = 0) = me^{-(kx)^2}$}\hbox{\strut $k = 2, m = 10$}} \\
    \hline
    \vtop{\hbox{\strut Non-linear Burgers:}\hbox{\strut $u_t + uu_x - vu_{xx} = 0$}} & $v = 0.001$ & \vtop{\hbox{\strut $x\in[-2,2]$}\hbox{\strut  $t\in[0,2]$}} & \vtop{\hbox{\strut $u(x,t = 0) = me^{-(kx)^2}$}\hbox{\strut $k = 2, m = 1$}} \\
\bottomrule
\end{tabular}
\label{problem-definitions}
\end{table*}

\section{Optimization Algorithms and Implementations}

\subsection{Neuroevolution by Evolution Strategies}

In this paper, we use two state-of-the-art probabilistic model-based neuroevolution algorithms, namely the Covariance Matrix Adaptation Evolution Strategy (CMA-ES) \cite{hansen2016cma} and exponential Natural Evolution Strategies (xNES) \cite{wierstra2014natural}, for training the PINN parameters. Both CMA-ES and xNES are population-based, gradient-free optimization procedures that are capable of approaching the global minimum of complex objective functions defined on continuous search spaces. These methods adapt their search update steps based on the rank-based fitness landscape characteristics of the objective function gleaned from sampled solutions.

As instantiations of information-geometric optimization procedures \cite{ollivier2017information}, CMA-ES and xNES iteratively update the mean vector and covariance of a multivariate Gaussian search distribution from which populations of candidate solutions are sampled and evaluated against the objective function. The spread of the search distribution allows a population to more effectively explore the space of possible solutions, thereby avoiding the propensity of point-based optimization techniques (e.g., SGD) of getting stuck in local minima. This makes CMA-ES and xNES well-suited for complex optimization problems with multiple local minima, such as in deep learning with PINN loss functions where near globally optimal solutions are sought. In particular, we consider an improvement of the xNES where its updates are augmented with the well-known Nesterov Accelerated Gradient (NAG) step \cite{muehlebach2019dynamical}. We refer to this modified version as xNES+NAG. The performance of the two neuroevolution algorithms are compared against variants of gradient descent popularly used today for training deep learning models. 

In particular, while hundreds of deep learning optimizers have been proposed and published, Schmidt et al. showed in their recent work on benchmarking deep learning optimizers that SGD remains the most popular by far in the literature ~\cite{schmidt2021descending}. In addition, their extensive experiments revealed that SGD had the best test set performance for their MLP-based benchmark problem. Hence, we have chosen SGD as a baseline gradient descent-based algorithm for comparison in this work. 

\subsection{Implementations} 

The JAX framework provides a library for deep learning research and development that allows users to flexibly define, train and run deep learning models. It does not require extensive code customization and has demonstrated tremendous performance improvements by leveraging hardware accelerators such as GPUs and TPUs. Hence, we have chosen to use JAX for implementation of all the benchmark problems and codes. Specifically, we utilize the SGD and CMA-ES implementations from the Optax and EvoJAX packages ~\cite{tang2022evojax}, respectively. The xNES+NAG algorithm has also been implemented by us using this framework.

To facilitate testing of other optimizers on the proposed benchmark problems, the PINN architecture and training loss definitions are wrapped within a simple black-box function that can take neural network weight parameters as input and return training loss as the equivalent fitness function evaluation. The xNES+NAG codes are provided as examples to illustrate the plug-and-play nature of the benchmark suite and it is hoped that additional optimizers will be added to the benchmark suite in the future. They are provided on GitHub (\url{https://github.com/nicholassung97/Neuroevolution-of-PINNs}). 

\begin{table*}[t]
    \footnotesize
    \caption{Detailed definitions of loss formulation for benchmark problems.}
    \begin{tabularx}{\textwidth}{@{} p{2cm}Xp{7cm} @{}}
        \hline
        \textbf{Problem} & \textbf{PDE Loss ($L_{PDE}$)} & \textbf{BC/IC Loss ($L_{BC}$ or $L_{IC}$)} \\
        \hline
        Convection-diffusion & $\frac{1}{n}\sum^{n}_{i=1}(v\hat{u}_x(x_i)-k\hat{u}_{xx}(x_i))^2$ & $\frac{1}{2}((u(x=0)-\hat{u}(x=0))^2 + (u(x=1)-\hat{u}(x=1))^2)$\\
        \hline
        Projectile motion & $\frac{1}{n}\sum^{n}_{i=1}((\hat{x}_{tt}(t_i))^2 + (\hat{y}_{tt}(t_i)+g)^2)$ & $(x(t=0)-\hat{x}(t=0))^2 + (y(t=0)-\hat{y}(t=0))^2 + (x_t(t=0)-\hat{x}_t(t=0))^2 + (y_t(t=0)-\hat{y}_t(t=0))^2$\\
        \hline
        Korteweg–De Vries (KdV) & $\frac{1}{n}\sum^{n}_{i=1}(\hat{u}_t(x_i,t_i) + v_1\hat{u}(x_i,t_i)\hat{u}_x(x_i,t_i) + v_2\hat{u}_{xxx}(x_i,t_i))^2$ & $\frac{1}{n}\sum^{n}_{i=1}(u(x_i,0)-\hat{u}(x_i,0))^2$\\
        \hline
        Linearized Burgers & $\frac{1}{n}\sum^{n}_{i=1}(\hat{u}_t(x_i,t_i) + v_1\hat{u}_x(x_i,t_i) - v_2\hat{u}_{xx}(x_i,t_i))^2$ & $\frac{1}{n}\sum^{n}_{i=1}(u(x_i,0)-\hat{u}(x_i,0))^2$\\
        \hline
        Non-linear Burgers & $\frac{1}{n}\sum^{n}_{i=1}(\hat{u}_t(x_i,t_i) + \hat{u}(x_i,t_i)\hat{u}_x(x_i,t_i) - v_1\hat{u}_{xx}(x_i,t_i))^2$ & $\frac{1}{n}\sum^{n}_{i=1}(u(x_i,0)-\hat{u}(x_i,0))^2$\\
        \hline
    \end{tabularx}
    \label{loss-functions}
\end{table*}

\section{Benchmark Problems}

In this paper, we propose five benchmark problems for PINNs which are representative of real-world phenomena. These differential equation problems were chosen to represent diverse phenomena across classical mechanics, heat and mass transfer, fluid dynamics and wave propagation (e.g. acoustics): 

\begin{itemize}
    \item \textbf{Steady state convection-diffusion equation} is an ordinary differential equation (ODE) that describes the final distribution of a scalar quantity in one spatial dimension in the presence of both convective transport and diffusion. It is commonly used to model transport phenomena in engineering and scientific applications. Using this equation, we simulate the temperature in a 1-D spatial domain. 
    \item \textbf{Projectile motion equation} is another set of ODEs that governs the motion of an object in a horizontal plane under the influence of gravity $g$ as described by the laws of classical mechanics. Using this equation, we simulate the projectile motion of an object released with a pre-determined initial velocity and angle of attack under a constant $g$.
    \item \textbf{Korteweg-De Vries (KdV) equation} is a partial differential equation (PDE) that describes the behaviour of weakly nonlinear, dispersive waves such as in shallow water or plasmas. The KdV equation is a simple, yet powerful model that captures the essential physics of these wave phenomena and has been widely used in a variety of applications, including fluid mechanics, plasma physics, and nonlinear optics. Using this equation, we simulate the collision of two waves of different magnitudes traveling from different locations. 
    \item \textbf{Linearized Burgers’ equation} is a PDE that describes the behaviour of viscous flow in one spatial dimension. It is a simplified version of the more general Burgers’ equation which is used to model the behaviour of fluid flow in the presence of shock waves and other complex phenomena. Using this equation, we simulate the propagation of a single waveform at constant velocity.
    \item \textbf{Non-linear Burgers’ equation} is a more complex and realistic version of the Burgers’ equation. For this problem, we similarly simulate the propagation of a single waveform, but with the formation of a shock front.   
\end{itemize}

Specific definitions of these benchmark problems, including exact PDE parameters and boundary/initial conditions, are in Table~\ref{problem-definitions}.

\section{Performance Metrics}

Three performance metrics were used to evaluate the effectiveness and efficiency of the optimization methods: training time, training loss, and prediction MSE.

\subsection{Training Time}
Training time is the amount of time it takes for the optimization method to train the PINN model. There are several factors that can affect the training time of the PINN model, including the sample size of the training data, the complexity of the physical laws being modelled, the number of network parameters, and the computational resources available for training. Detailed descriptions of these parameters are provided in subsequent sections to facilitate benchmarking and comparison across optimization methods. Based on our results, we choose to evaluate the training performance for the PINN benchmark problems at the 60s and 180s time cut-off.

\subsection{Training Loss}
Training loss is a measure of the difference between the predictions made by the PINNs and the known physical laws governing the system being modelled during training, and indicates the degree of violation of the governing physics. It is calculated by summing the difference between the predicted outputs and the true values at the initial and boundary conditions, as well as the residual of the partial differential equation (PDE) representing the physical laws, as defined in Equation \ref{eq:pde_ibc_eqn}. For this work, weights of the loss terms, which affects the relative importance of each term in the loss function, is kept at one. The training loss function ($L_{PINN}$) is defined in Table~\ref{loss-functions} for each benchmark problem. 

Importantly, this same training loss is used to evaluate the network’s performance during the training process. Fundamentally, the goal of any optimization method is to train the PINN models to minimize the training loss such that the PINNs’ predictions are as close as possible to the real-world physical system.

\subsection{Prediction Mean Squared Error}

The prediction mean squared error (MSE) refers to the error between the predicted solution and the ground truth solution. It is calculated by taking the average of the squared difference between the model’s predictions and the true target values. The prediction MSE is important because it provides validation of the model's performance that is independent of the training loss. A lower prediction MSE against the ground truth solution indicates better performance of the optimization method in training the PINN model.

To calculate the prediction MSE, we need the ground truth solution for each benchmark problem. For the convection-diffusion and the projectile motion problems, there is an analytical ground truth solution. In contrast, the ground truth solution for the Korteweg-De Vries (KdV), linearized Burgers, and non-linear Burgers’ problems need to be obtained by numerical simulation. They are obtained by utilizing a finite volume scheme for spatial terms, and a second order Runge-Kutta method for temporal terms. To ensure convective stability, the convection terms are approximated by second order dispersion-relation preserving finite volume scheme in conjunction with a universal limiter \cite{chiu2018improved,leonard1991ultimate}, while remaining spatial terms are approximated by central difference. All ground truth solutions are provided on GitHub (\url{https://github.com/nicholassung97/Neuroevolution-of-PINNs})

\begin{table*}[htbp]
\centering
\small 
\caption{Configurations of PINNs and optimization algorithms used.}
\begin{tabularx}{\textwidth}{@{} X X X X X X @{}}
\toprule
\textbf{Problem} & \textbf{PINN architecture$^{*}$} & \textbf{Collocation points (incl. IC/BC points)} & \textbf{Population size, initial $\sigma$ [CMA-ES]} & \textbf{Population size, learning rate, initial $\sigma$, momentum [xNES+NAG]} & \textbf{PDE + IC/BC mini-batch, learning rate [SGD]} \\
\midrule
Convection-diffusion & $(x)-10-10-10-(\hat{u})$, network weights = 250 & 10,000 (2) & 80, 5e-2 & 100, 1e-2, 1e-3, 0.99 & 100 + 2, 1e-3 \\
Projectile motion & $(t)-8-8-[8-(\hat{x}), 8-(\hat{y})]$, network weights = 240 & 10,000 (1) & 80, 1e-3 & 50, 1e-3, 1e-3, 0.99 & 100 + 1, 1e-3 \\
Korteweg--De Vries (KdV) & $(x,t)-8-8-8-8-(\hat{u})$, network weights = 240 & 15,477 (77) & 50, 5e-2 & 100, 1e-2, 1e-3, 0.9 & 100 + 5, 1e-1 \\
Linearized Burgers & $(x,t)-10-10-10-(\hat{u})$, network weights = 260 & 38,793 (193) & 50, 1e-2 & 100, 1e-2, 1e-3, 0.9 & 100 + 5, 1e-2 \\
Non-linear Burgers & $(x,t)-8-8-8-(\hat{u})$, network weights = 240 & 25,929 (129) & 100, 1e-2 & 50, 1e-3, 1e-3, 0.99 & 100 + 5, 1e-1 \\
\bottomrule
\end{tabularx}

\begin{description}[leftmargin=0pt]

\item The numbers between input and output represent the number of nodes in hidden layers of the PINN. For example, $(x)-10-10-10-(\hat{u})$ indicates a neural net with single input $x$, followed by 3 hidden layers with 10 nodes each, and a single output $\hat{u}$. All hidden layers, except the final hidden layer, include a bias term and use \textit{tanh} activation function. The final hidden layer uses \textit{linear} activation function and does not include a bias term.
\end{description}
\label{tab04}
\end{table*}

\begin{figure*}[htbp]
\begin{center}
\centerline{\includegraphics[width=1\linewidth]{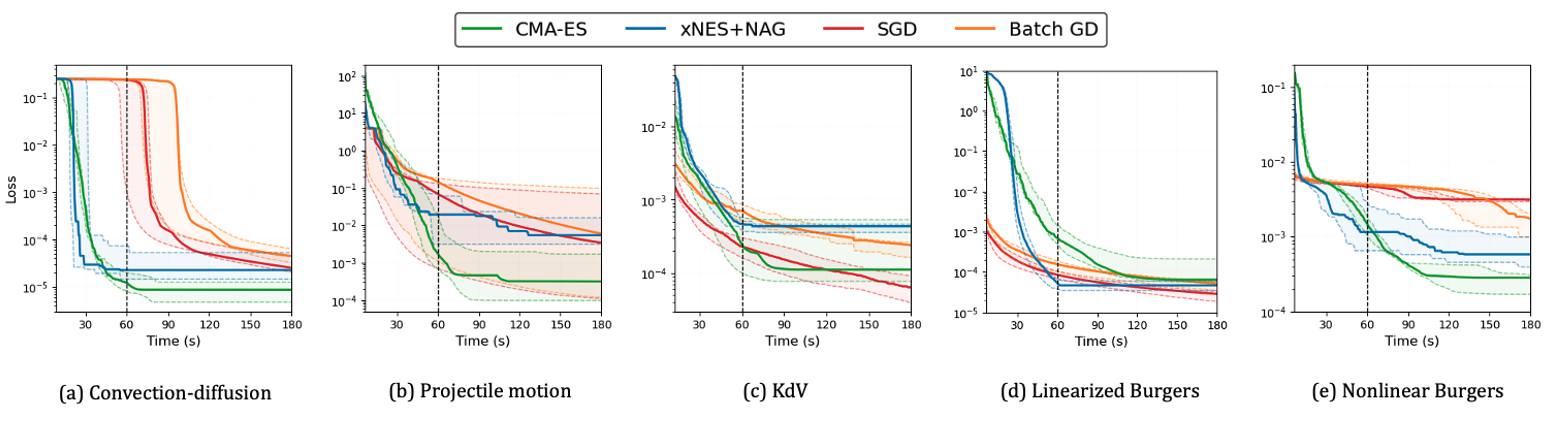}}
\caption{Loss convergence plots monitored up to 180s.}
\label{fig:loss-plots}
\end{center}
\vspace*{-8mm}
\end{figure*}

\section{Experimental Study}

This section presents results from using neuroevolution to solve the five benchmark problems. Critically, the results indicate that EAs, such as CMA-ES, can be as effective at solving PINN problems as SGD.

\subsection{Configurations of PINNs and Optimization Algorithms}
A preliminary investigation was conducted to explore the relationship between neural network size and model performance by varying the number of neurons in the hidden layers. An adequate size was selected for each problem to ensure good approximations of the solutions to the differential equations. 

Additionally, we carried out a hyperparameter search to identify the best settings for all optimizers. For CMA-ES, we tested a range of population sizes (20, 50, 80, 100) and initial standard deviations of a zero-mean Gaussian distribution model (0.001, 0.005, 0.01, 0.05, 0.1, 0.5). For xNES+NAG, we tested a range of population sizes (20, 50, 80, 100), learning rates (0.001,0.01), initial standard deviations (0.001, 0.01), and momentum (0.9, 0.99, 0.999). For SGD, we tested a range of learning rates (0.001, 0.005, 0.01, 0.05, 0.1, 0.5) and minibatch sampling size of collocation points for both interior domain (10, 100, 1000), and boundary/initial points (5, 50) for benchmark problems 3-5. Xavier initialization was also used with standard parameters. In prior PINN work, other authors have reported that gradient-based optimizers like full-batch L-BFGS can improve the PINN model performance \cite{raissi2019physics}. Hence, in the context of this work, where the neural network parameters are relatively fewer and memory scalability is not an issue, we report results from SGD optimized with both the optimal mini-batch size from hyper-parameter search and without mini-batching (similar to batch gradient descent (batch GD)) for completeness. 

For reproducibility, a full description of the PINN configurations and best settings for tested optimization algorithms as used for each benchmark problem can be found in Table~\ref{tab04}.

\subsection{Baseline Results}
For the sake of consistency, all results are obtained from runs on a workstation with an Intel Xeon W-2275 Processor (19.25M Cache, 3.30 GHz, 14 cores) CPU and a NVIDIA GeForce RTX 3090 GPU. 

\begin{figure*}[htbp]
\begin{center}
\centerline{\includegraphics[width=.7\linewidth]{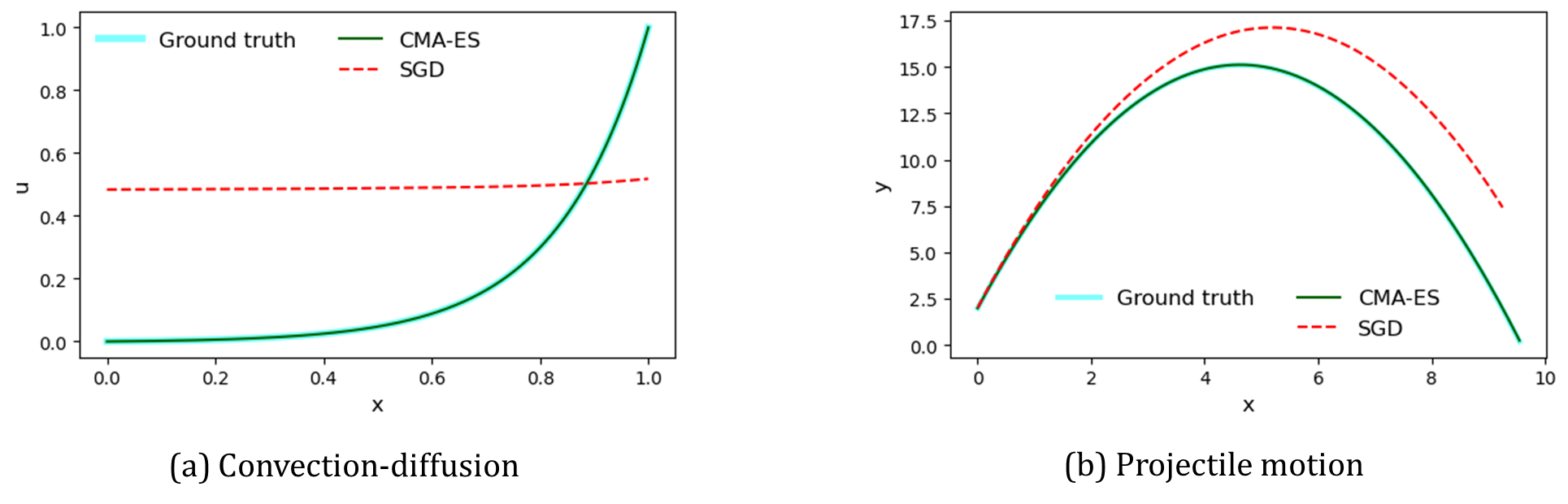}}
\caption{Plots of solutions obtained for (a) convection-diffusion and (b) projectile motion after 60 sec of training.}
\label{fig:solution-a-b}
\end{center}
\vspace*{-5mm}
\end{figure*}

\begin{figure*}[htbp]
\begin{center}
\centerline{\includegraphics[width=0.85\linewidth]{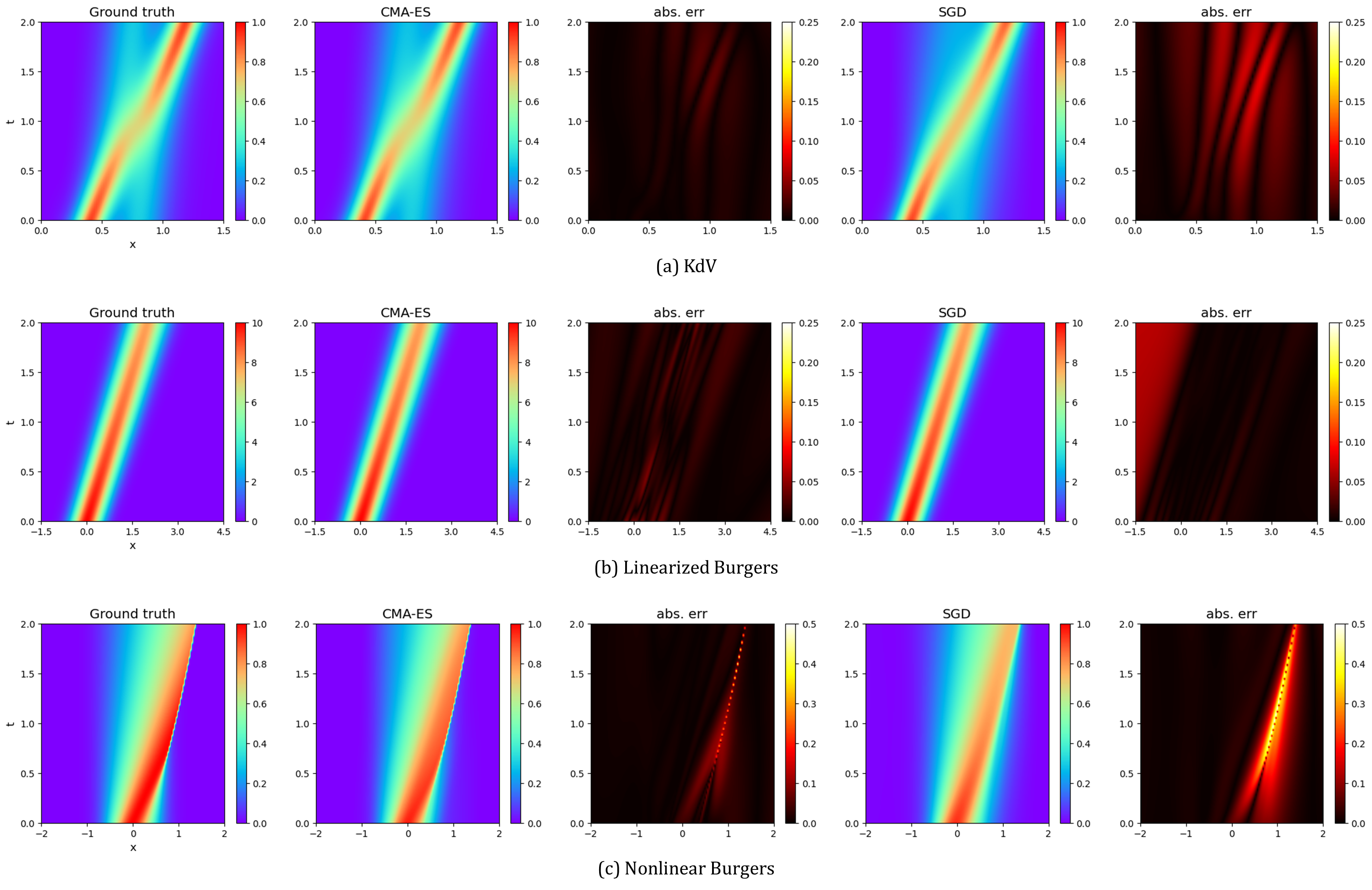}}
\caption{Plots of solutions obtained for (a) Korteweg–De Vries (KdV), (b) linearized Burgers, and (c) non-linear Burgers after 60 sec of training.}
\label{fig:solution-c-d-e}
\end{center}
\vspace*{-10mm}
\end{figure*}

\subsubsection{Convergence Behavior}

In this study, five independent runs of each of the optimization methods were performed and the convergence was monitored up to 180s. The convergence trends are plotted against time for all 5 benchmark problems in Fig.~\ref{fig:loss-plots}. The bold lines on the plot represent the median convergence path, and the shaded areas indicate the range of values from the minimum to the maximum convergence path across the five runs. 

Fig.~\ref{fig:loss-plots} shows that the two neuroevolution algorithms generally converge to the minimum much faster than SGD. Interestingly, the plots of loss convergence for the optimizers show an intersection, indicating that the various optimizers perform differently at different stages of the optimization process. This suggests possible future hybridizations may be able to leverage the best of both worlds. In addition, the results suggest that SGD typically converges more slowly than CMA-ES and xNES+NAG, potentially due to the neuroevolutionary algorithms being better at solving complex PINN optimization landscapes with many local optima. In particular, the neuroevolutionary algorithms use multiple samples to explore the search space, and can therefore navigate across a complex landscape with local optima more effectively than gradient descent \cite{garipov2018loss}. SGD may also be less effective at the start as it can become biased towards local optima near its initialization as described in Section~\ref{loss-initialization}, thereby delaying the onset of convergence.

\subsubsection{Training Loss and Prediction MSE} \label{CMA-ES-vs-SGD}

In a practical context, it is important that PINN training be completed in a reasonable amount of time. Hence, we further compare the training loss and prediction MSE attained after 60 seconds of training. The best MSE predictions are highlighted in bold in Table~\ref{tab: all-error}. In all cases, the prediction MSE is lowest for a neuroevolution algorithm, although xNES+NAG and CMA-ES out-perform each other on different problems.

\begin{table}[htpb]
\footnotesize
\caption{Training loss and prediction MSE against ground truth solution at 60 seconds. The best MSE, marked in bold, is always achieved by a neuroevolution algorithm.}
\begin{center}
\begin{tabular}{llll}\toprule
\textbf{Problem} & \textbf{Optimizer} & \textbf{Training Loss} & \textbf{Prediction MSE} \\
\hline
\multirow{4}{*}{\centering Convection-diffusion} & {\bf{CMA-ES}} & {$1.00 \times 10^{-5}$} & {$\textbf{6.38} \times \bf{10^{-9}}$} \\
\cline{2-4}
{} & {xNES+NAG} & {$5.05 \times 10^{-5}$} & {$7.44 \times 10^{-7}$} \\
\cline{2-4}
{} & {SGD} & {$2.40 \times 10^{-1}$} & {$1.59 \times 10^{-1}$} \\
\cline{2-4}
{} & {Batch GD} & {$2.44 \times 10^{-1}$} & {$1.64 \times 10^{-1}$} \\
\hline
\multirow{4}{*}{\centering Projectile motion} & {\bf{CMA-ES}} & {$6.73 \times 10^{-4}$} & {$\textbf{3.96} \times \bf{10^{-4}}$} \\
\cline{2-4}
{} & {xNES+NAG} & {$1.83 \times 10^{-2}$} & {$2.89 \times 10^{-3}$} \\
\cline{2-4}
{} & {SGD} & {$1.27 \times 10^{-1}$} & {$5.50 \times 10^{0}$} \\
\cline{2-4}
{} & {Batch GD} & {$1.79 \times 10^{-1}$} & {$6.24 \times 10^{0}$} \\
\hline
\multirow{4}{*}{\centering Korteweg–De Vries (KdV)} & {\bf{CMA-ES}} & {$1.02 \times 10^{-4}$} & {$\textbf{7.57} \times \bf{10^{-5}}$} \\
\cline{2-4}
{} & {xNES+NAG} & {$3.89 \times 10^{-4}$} & {$7.58 \times 10^{-4}$} \\
\cline{2-4}
{} & {SGD} & {$3.12 \times 10^{-4}$} & {$7.33 \times 10^{-4}$} \\
\cline{2-4}
{} & {Batch GD} & {$8.73 \times 10^{-4}$} & {$8.53 \times 10^{-4}$} \\
\hline
\multirow{4}{*}{\centering Linearized Burgers} & {CMA-ES} & {$1.07 \times 10^{-3}$} & {$9.83 \times 10^{-5}$} \\
\cline{2-4}
{} & {\bf{xNES+NAG}} & {$6.80 \times 10^{-5}$} & {$\textbf{4.54} \times \bf{10^{-5}}$} \\
\cline{2-4}
{} & {SGD} & {$8.50 \times 10^{-5}$} & {$4.84 \times 10^{-4}$} \\
\cline{2-4}
{} & {Batch GD} & {$1.50 \times 10^{-4}$} & {$7.13 \times 10^{-4}$} \\
\hline
\multirow{4}{*}{\centering Non-Linear Burgers} & {\bf{CMA-ES}} & {$1.22 \times 10^{-3}$} & {$\textbf{6.82} \times \bf{10^{-4}}$} \\
\cline{2-4}
{} & {xNES+NAG} & {$2.96 \times 10^{-3}$} & {$3.23 \times 10^{-3}$} \\
\cline{2-4}
{} & {SGD} & {$4.84 \times 10^{-3}$} & {$5.92 \times 10^{-3}$} \\
\cline{2-4}
{} & {Batch GD} & {$5.29 \times 10^{-3}$} & {$6.88 \times 10^{-3}$} \\
\bottomrule
\end{tabular}
\label{tab: all-error}
\end{center}
\end{table}

The results show that CMA-ES and xNES+NAG outperform SGD on training loss for four out of the five benchmarks each. Importantly, CMA-ES generates the lowest prediction MSE for four of the five benchmark problems, while xNES+NAG has the lowest prediction MSE for the other benchmark problem. 

To visualize the results, we further plot the solutions generated by CMA-ES and SGD after 60 seconds of training (Fig.~\ref{fig:solution-a-b} and Fig.~\ref{fig:solution-c-d-e}). We chose CMA-ES for visualization as it yielded the lowest MSE on four of the five benchmark problems, as opposed to xNES+NAG's one. For the convection-diffusion (Fig.~\ref{fig:solution-a-b}a), projectile motion (Fig.~\ref{fig:solution-a-b}b), KdV (Fig.~\ref{fig:solution-c-d-e}a), and non-linear Burgers’ (Fig.~\ref{fig:solution-c-d-e}c), the solutions generated by CMA-ES are noticeably more similar to the ground truth than SGD. The absolute error plots in Fig.~\ref{fig:solution-c-d-e} clearly illustrate a smaller error, and a better match to ground truth, for CMA-ES. However, the solution for both optimizers are visually similar to the simulated solution for the linearized Burgers’ problem (Fig.~\ref{fig:solution-c-d-e}b).

\section{Conclusion}

We examined the effectiveness of SGD variants in optimizing PINNs in comparison to DNNs. Our analysis revealed that the presence of spurious local minima in proximity to the initialization point may pose challenges to the effectiveness of optimizer choices like SGD. In light of these findings, we established the necessity for alternative optimization techniques, such as neuroevolution, which exhibit greater global exploration capacity to avoid local minima traps. 

To facilitate evaluation of different optimizers, we also proposed five PINN benchmark problems. We demonstrated that evolutionary algorithms, as exemplified by probabilistic model-based CMA-ES and xNES, have the potential to be more effective than SGD. Importantly, neuroevolution algorithms generate the lowest prediction MSE against the ground truth for all five benchmark problems after 60 seconds of training. While these results do not conclusively guarantee that EAs will surpass SGD across all PINN problems, they are an intriguing hint as to the potential for EAs in this area. At the least, it suggests that the exploration and development of alternative optimization algorithms, including EAs, for the complex task of optimizing PINNs is warranted for real-world scientific and engineering problems. Hybridizing global search with local gradient signals is yet another promising direction in this regard.

\bibliographystyle{unsrt}
\bibliography{template}  






\end{document}